\documentclass[11pt,a4paper]{article}
\usepackage[hyperref]{emnlp2018}
\usepackage{times}
\usepackage{latexsym}
\usepackage{paralist}
\usepackage{multirow}
\usepackage{graphicx}
\usepackage{amsmath, amssymb, verbatim}
\usepackage{url}

\usepackage{amsfonts}
\usepackage{makecell}
\usepackage{hhline}
\usepackage{multirow, rotating}
\usepackage{bbding}
\usepackage{algorithm, algorithmic, amsthm}
\usepackage{color}
\usepackage{float}
\usepackage{subfigure}
\usepackage{todonotes}
\usepackage{enumitem}

\aclfinalcopy 

\newcommand{\nop}[1]{}

\title{Discriminative Deep Dyna-Q: Robust Planning \\ for Dialogue Policy Learning}

\author{Shang-Yu Su$^{\star}$\quad Xiujun Li$^{\dagger}$\quad Jianfeng Gao$^{\dagger}$\quad Jingjing Liu$^{\dagger}$\quad Yun-Nung Chen$^{\star}$  \\
  $^{\dagger}$Microsoft Research, Redmond, WA, USA \\
  $^{\star}$National Taiwan University, Taipei, Taiwan \\
  {\tt $^\star$f05921117@ntu.edu.tw\quad $^\star$y.v.chen@ieee.org} \\
  {\tt $^\dagger$\{xiul,jfgao,jingjl\}@microsoft.com}
}

\date{}

\begin{document}
\maketitle

\begin{abstract}
This paper presents a Discriminative Deep Dyna-Q (D3Q) approach to improving the effectiveness and robustness of Deep Dyna-Q (DDQ), a recently proposed framework that extends the Dyna-Q algorithm to integrate planning for task-completion dialogue policy learning. To obviate DDQ's high dependency on the quality of simulated experiences, we incorporate an RNN-based discriminator in D3Q to differentiate simulated experience from real user experience in order to control the quality of training data. Experiments show that D3Q significantly outperforms DDQ by controlling the quality of simulated experience used for planning. The effectiveness and robustness of D3Q is further demonstrated in a domain extension setting, where the agent's capability of adapting to a changing environment is tested.\footnote{The source code is available at \url{https://github.com/MiuLab/D3Q}.}
\end{abstract}

\section{Introduction}
There are many virtual assistants commercially available today, such as Apple's Siri, Google's Home, Microsoft's Cortana, and Amazon's Echo. With a well-designed dialogue system as an intelligent assistant, people can accomplish tasks via natural language interactions. Recent advance in deep learning has also inspired many studies in neural dialogue systems~\cite{wen2017network, bordes2017learning, dhingra2017towards, li2017end}. 

A key component in such task-completion dialogue systems is \emph{dialogue policy}, which is often formulated as a reinforcement learning (RL) problem~\cite{levin1997learning,young2013pomdp}. However, learning dialogue policy via RL from the scratch in real-world systems is very challenging, due to the inevitable dependency on the environment from which a learner acquires knowledge and receives rewards. In a dialogue scenario, real users act as the environment in the RL framework, and the system communicates with real users constantly to learn dialogue policy. Such process is very time-consuming and expensive for online learning.

One plausible strategy is to leverage user simulators trained on human conversational data~\cite{schatzmann2007agenda,li2016user}, which allows the agent to learn dialogue policy by interacting with the simulator instead of real users. The user simulator can provide infinite simulated experiences without additional cost, and
the trained system can be deployed and then fine-tuned through interactions with real users~\cite{su2016continuously, lipton2016efficient, zhao2016towards, williams2017hybrid, dhingra2017towards, li2017end, liu2017iterative, peng2017composite, budzianowski2017sub, peng2017adversarial, tang2018subgoal}.  


However, due to the complexity of real conversations and biases in the design of user simulators, there always exists the discrepancy between real users and simulated users. Furthermore, to the best of our knowledge, there is no universally accepted metric for evaluating user simulators for dialogue purpose~\cite{pietquin2013survey}. 
Therefore, it remains controversial whether training task-completion dialogue agent via simulated users is a valid and effective approach.

A previous study, called Deep Dyna-Q (DDQ)~\cite{peng2018integrating}, proposed a new strategy to learn dialogue policies with real users by combining the Dyna-Q framework~\cite{sutton1990integrated} with deep learning models. 
This framework incorporates a learnable environment model (\emph{world model}) into the dialogue policy learning pipeline, which simulates dynamics of the environment and generates simulated user behaviors to supplement the limited amount of real user experience. In DDQ, real user experiences play two pivotal roles: 1) directly improve the dialogue policy via RL; 2) improve the world model via supervised learning to make it behave more human-like. The former is referred to as \textit{direct reinforcement learning}, and the latter \textit{world model learning}. Respectively, the policy model is trained via real experiences collected by interacting with real users (\textit{direct reinforcement learning}), and simulated experiences collected by interacting with the learned world model (\textit{planning} or \textit{indirect reinforcement learning}).

However, the effectiveness of DDQ depends upon the quality of simulated experiences used in planning. As pointed out in \cite{peng2018integrating}, although at the early stages of dialogue training it is helpful to perform planning aggressively with large amounts of simulated experiences regardless their quality, in the late stages when the dialogue agent has been significantly improved, low-quality simulated experiences often hurt the performance badly. Since there is no established method of evaluating the world model which generates simulated experiences, \citet{peng2018integrating} resorts to heuristics to mitigate the negative impact of low-quality simulated experiments, e.g., reducing the planning steps in the late stage of training. These heuristics need to be tweaked empirically, thus limit DDQ's applicability in real-world tasks.


\begin{figure}[t]
\centering 
\includegraphics[width=\linewidth]{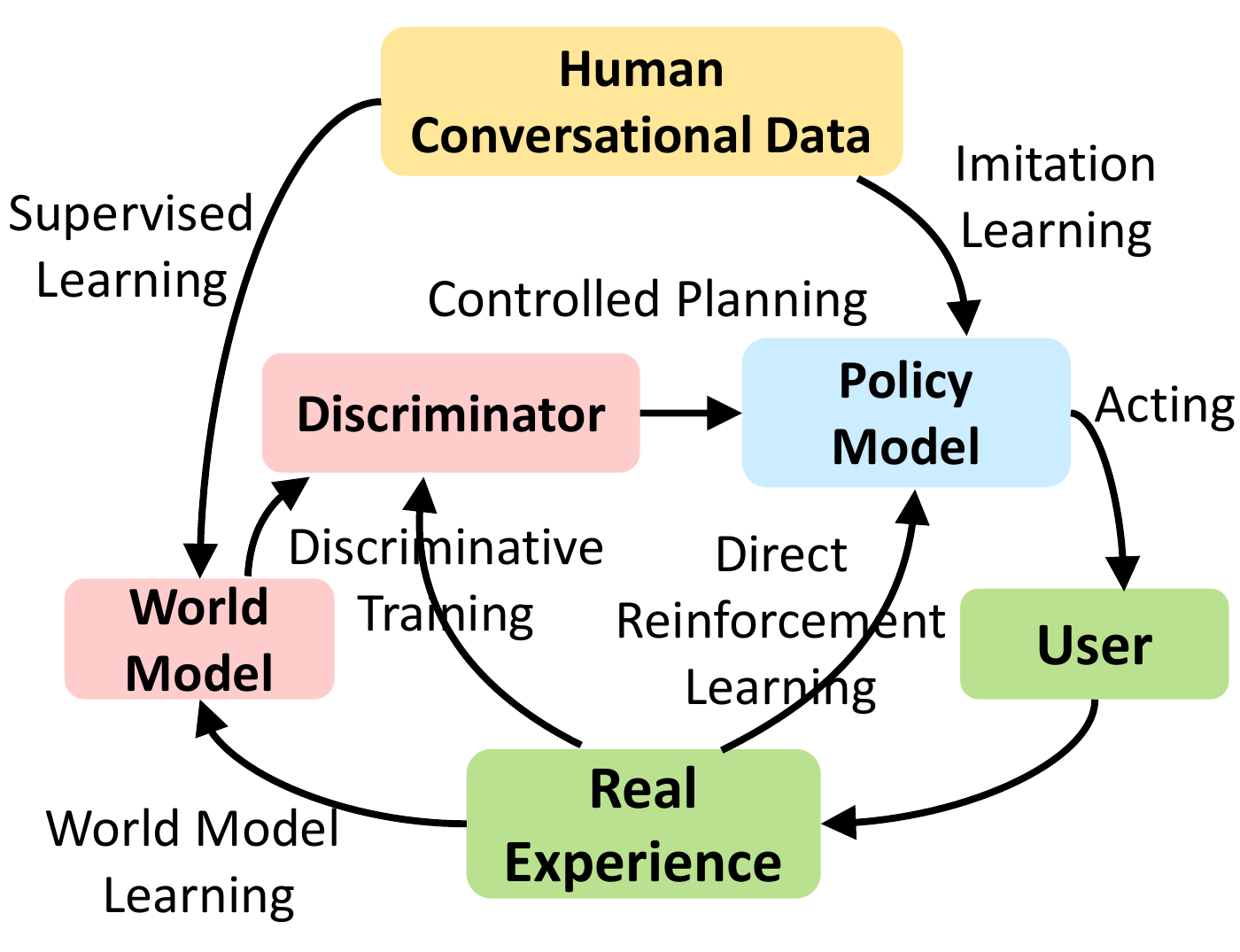}
\vspace{-5mm}
\caption{Proposed D3Q for dialogue policy learning.} 
\label{fig:d3q} 
\end{figure}

To improve the effectiveness of planning without relying on heuristics, this paper proposes Discriminative Deep Dyna-Q (D3Q), a new framework inspired by generative adversarial network (GAN) that incorporates a discriminator into the planning process. The discriminator is trained to differentiate simulated experiences from real user experiences. As illustrated in Figure~\ref{fig:d3q}, all simulated experiences generated by the world model need to be \textit{judged} by the discriminator, only the high-quality ones, which cannot be easily detected by the discriminator as being simulated, are used for planning. During the course of dialogue training, both the world model and discriminator are refined using the real experiences. So, the quality threshold held by the discriminator goes up with the world model and dialogue agent, especially in the late stage of training. 

By employing the world model for planning and a discriminator for controlling the quality of simulated experiences, the proposed D3Q framework can be viewed as a model-based RL approach, which is generic and can be easily extended to other RL problems. In contrast, most model-based RL methods~\cite{tamar2016value,silver2016predictron,gu2016continuous,racaniere2017imagination} are developed for simulation-based, synthetic problems (e.g., games), not for real-world problems. In summary, our main contributions in this work are two-fold:
\begin{compactitem}
\item The proposed Discriminative Deep Dyna-Q approach is capable of controlling the quality of simulated experiences generated by the world model in the planning phase, which enables effective and robust dialogue policy learning.
\item The proposed model is verified by experiments including simulation, human evaluation, and domain-extension settings, where all results show better sample efficiency over the DDQ baselines.
\end{compactitem}

\section{Discriminative Deep Dyna-Q (D3Q)}
\begin{figure}[t]
\centering 
\includegraphics[width=\linewidth]{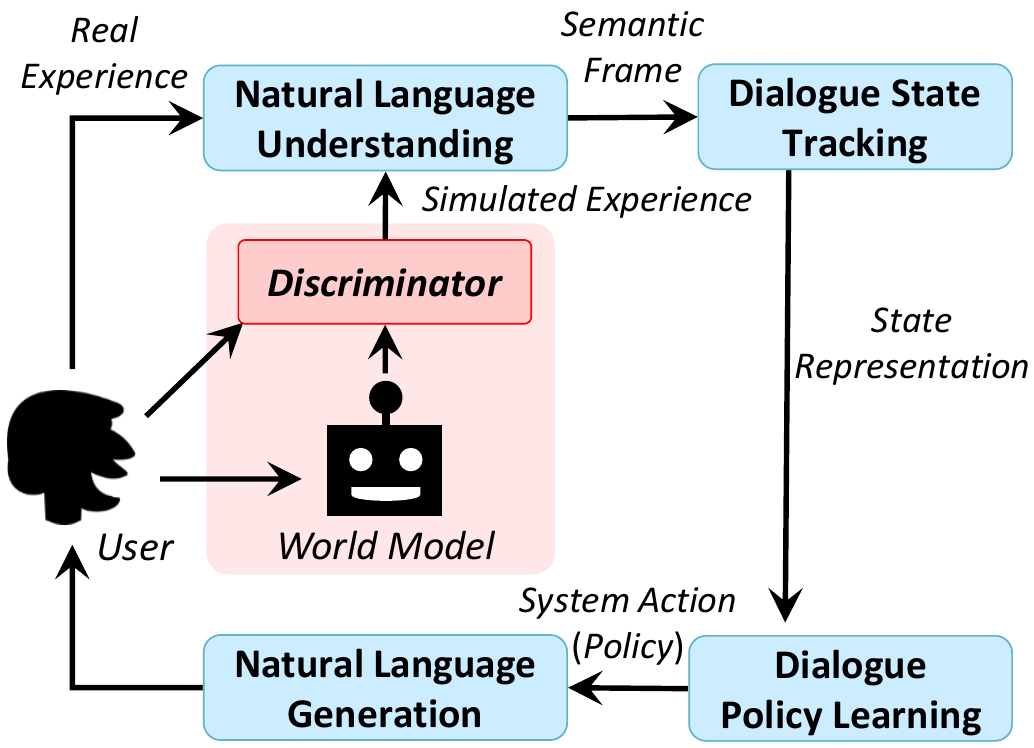}
\vspace{-3mm}
\caption{Illustration of the proposed D3Q dialogue system framework.} 
\label{fig:framework} 
\end{figure}

As illustrated in Figure~\ref{fig:framework}, the D3Q framework consists of six modules:
(1) an LSTM-based natural language understanding (NLU) module~\cite{hakkani-tur2016multi} for identifying user intents and extracting associated slots;
(2) a state tracker~\cite{mrkvsic2017neural} for tracking dialogue states; 
(3) a dialogue policy that selects next action based on the current state; 
(4) a model-based natural language generation (NLG) module for generating natural language response~\cite{wen2015semantically};
(5) a world model for generating simulated user actions and simulated rewards; 
and (6) an RNN-based discriminator for controlling the quality of simulated experience.
Note that the controlled planning phase is realized through the world model and the discriminator, which are not included in traditional framework of dialogue systems.

Figure~\ref{fig:d3q} illustrates the whole process: starting with an initial dialogue policy and an initial world model (both are trained with pre-collected human conversational data), D3Q training consists of four stages:
(1) \emph{direct reinforcement learning}: the agent interacts with real users, collects real experiences and improves dialogue policy;
(2) \emph{world model learning}: the world model is learned and refined using real experience;
(3) \emph{discriminator learning}: the discriminator is learned and refined to differentiate simulated experience from real experience;
and (4) \emph{controlled planning}: the agent improves the dialogue policy using the high-quality simulated experience generated by the world model and the discriminator.

\subsection{Direct Reinforcement Learning}
In this stage, we use the vanilla deep Q-network (DQN) method~\cite{mnih2015human} to learn the dialogue policy based on real experience. We consider task-completion dialogue as a Markov Decision Process (MDP), where the agent interacts with a user through a sequence of actions to accomplish a specific user goal.

At each step, the agent observes the dialogue state $s$, and chooses an action $a$ to execute, using an $\epsilon$-greedy policy that selects a random action with probability $\epsilon$ or otherwise follows the greedy policy $a=\text{argmax}_{a'} Q(s,a';\theta_{Q})$. $Q(s,a;\theta_{Q})$ which is the approximated value function, implemented as a Multi-Layer Perceptron (MLP) parameterized by $\theta_{Q}$. The agent then receives reward\nop{\footnote{In the task-completion dialogue scenario, reward is defined to measure the degree of success of a dialogue. In our experiment, for example, success corresponds to a reward of $80$, failure to a reward of $-40$, and the agent receives a reward of $-1$ at each turn so as to encourage shorter dialogues.}} $r$, observes next user response, and updates the state to $s'$. Finally, we store the experience tuple $(s, a, r, s')$ in the replay buffer $B^u$. This cycle continues until the dialogue terminates. 

We improve the value function $Q(s,a;\theta_{Q})$ by adjusting $\theta_{Q}$ to minimize the mean-squared loss function as follows:
\begin{eqnarray}
\label{eq:dqn}
\mathcal{L}(\theta_{Q})&=&\mathbb{E}_{(s,a,r,s')\sim B^u}[(y_i - Q(s, a;\theta_{Q}))^2], \nonumber \\
 y_i&=&r +\gamma \max_{a'}Q'(s', a';\theta_{Q'}),
\end{eqnarray}
where $\gamma \in [0,1]$ is a discount factor, and $Q'(.)$ is the target value function that is only periodically updated (i.e., fixed-target). The dialogue policy can be optimized through $\nabla_{\theta_{Q}}\mathcal{L}(\theta_{Q})$ by mini-batch deep Q-learning.

\subsection{World Model Learning}

\begin{figure*}[t]
\centering
\includegraphics[width=\linewidth]{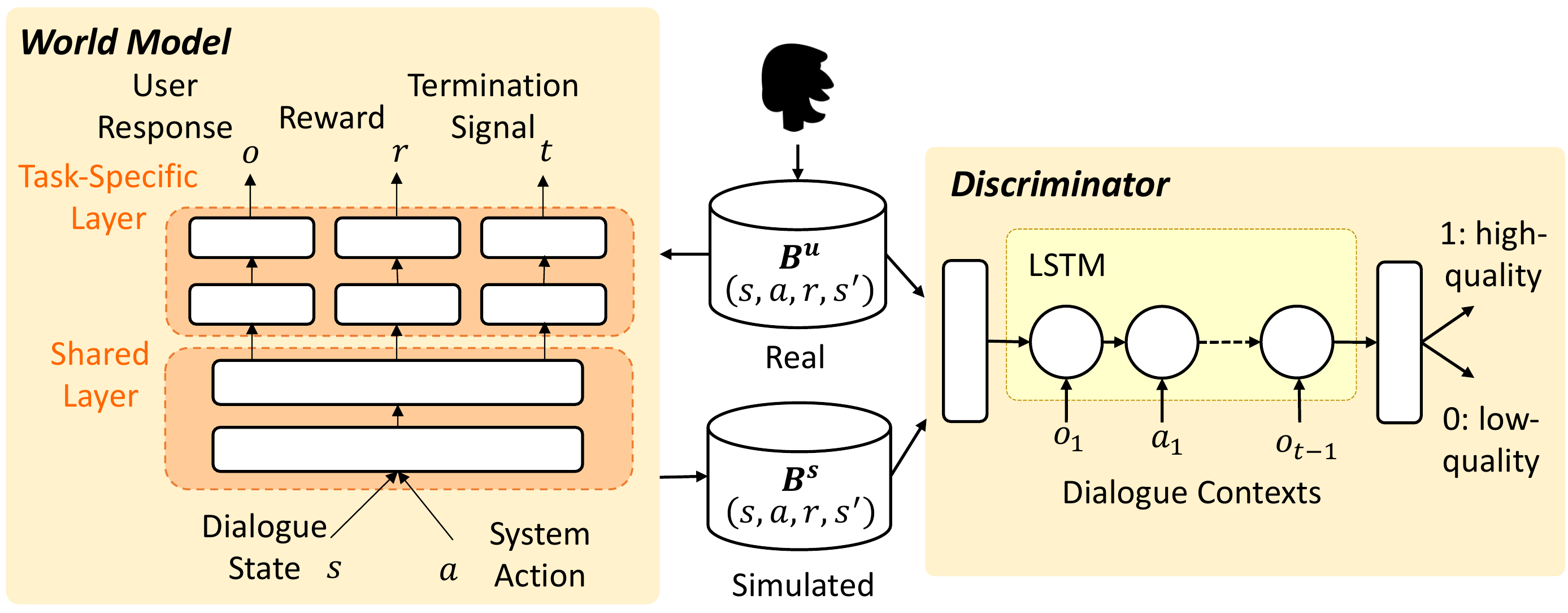}
\vspace{-4mm}
\caption{The model architectures of the world model and the discriminator for controlled planning.}
\label{fig:planner_arch}
\end{figure*}

To enable planning, we use a world model to generate simulated experiences that can be used to improve dialogue policy. In each turn of a dialogue, the world model takes the current dialogue state $s$ and the last system action $a$ (represented as an one-hot vector) as the input, and generates the corresponding user response $o$, reward $r$, and a binary variable $t$ (indicating if the dialogue terminates). The world model $G(s, a; \theta_{G})$ is trained using a multi-task deep neural network~\cite{liu2015representation} to generate the simulated experiences. The model contains two classification tasks for simulating user responses $o$ and generating terminal signals $t$, and one regression task for generating the reward $r$. The lower encoding layers are shared across all three tasks, while the upper layers are task-specific. $G(s,a;\theta_{G})$ is optimized to mimic human behaviors by leveraging real experiences in the replay buffer $B^u$. The model architecture is illustrated in the left part of Figure~\ref{fig:planner_arch}.
\begin{eqnarray*}
h &=& \tanh(W_{h} (s, a) + b_h),\\
r &=& W_r h + b_r, \\
o &=& \texttt{softmax} (W_a h + b_a), \\
t &=& \texttt{sigmoid} (W_t h + b_t),
\end{eqnarray*}
where $(s,a)$ is the concatenation of $s$ and $a$, and all $W$ and $b$ are weight matrices and bias vectors, respectively.

\subsection{Discriminator Learning}

The discriminator, denoted by $D$, is used to differentiate simulated experience from real experience. $D$ is a neural network model with its architecture illustrated in the right part of Figure \ref{fig:planner_arch}. $D$ employs an LSTM to encode a dialogue as a feature vector, and a Multi-Layer Perceptron (MLP) to map the vector to a probability indicating whether the dialogue looks like being generated by real users.

$D$ is trained using the simulated experience generated by the world model $G$ and the collected real experience $x$. We use the objective function as
\begin{equation}
\label{eq:discriminator_objective}
\mathbb{E}_{real}[\log D(x)] + \mathbb{E}_{simu}[\log(1-D(G(.)))].
\end{equation}
Practically, we use the mini-batch training and the objective function can be rewritten as
\begin{equation}
\label{eq:batch_discriminator_objective}
\frac{1}{m} \sum^{m}_{i=1}[\log D(x^{(i)})+\log(1-D(G(.)^{(i)}))],
\end{equation}
where $m$ represents the batch size.

\subsection{Controlled Planning}
\label{subsec:controlled_planning}
In this stage, we apply the world model $G$ and the discriminator $D$ to generate high-quality simulated experience to improve dialogue policy. 
The D3Q method uses three replay buffers, $B^u$ for storing real experience, $B^s$ for simulated experience generated by $G$, and $B^h$ for high-quality simulated experience generated by $G$ and $D$. Learning and planning are implemented by the same DQN algorithm, operating on real experience in $B^u$ for learning and on simulated experience in $B^h$ for planning. Here we only describe how the high-quality simulated experience is generated.

At the beginning of each dialogue session, we uniformly draw a user goal $(C,R)$~\cite{schatzmann2007agenda}, where $C$ is a set of constraints and $R$ is a set of requests. For example, in movie-ticket booking dialogue, \emph{constraints} are the slots with specified values, such as the name, the date of the movie and the number of tickets to buy. And \emph{requests} can contain slots which the user plans to acquire the values for, such as the start time of the movie. The first user action $o_1$ can be either a \texttt{request} or an \texttt{inform} dialogue act. A request dialogue act consists of a request slot, multiple constraint slots and the corresponding values, uniformly sampled from $R$ and $C$. For example, \texttt{request(theater; moviename=avergers3)}. An inform dialogue act contains constraint-slots only. Semantic frames can also be transformed into natural language via NLG component, e.g., ``\textit{which theater will play the movie avergers3?}''

For each dialogue episode with a sampled user goal, the agent interacts with world model $G(s, a; \theta_{G})$ to generate a simulated dialogue session, which is a sequence of simulated experience tuples $(s,a,r,s')$. We always store the $G$-generated session in $B^s$, but only store it in $B^h$ if it is selected by discriminator $D$. We repeat the process until $K$ simulated dialogue sessions are added in $B^h$, where $K$ is a pre-defined planning step size. This can be viewed as a sampling process. In theory if the world model $G$ is not well-trained this process could take forever to generate $K$ high-quality samples accepted by $D$. Fortunately, this never happened in our experiments because $D$ is trained using the simulated experience generated by $G$ and $D$ is updated whenever $G$ is refined.

Now, we compare controlled planning in D3Q with the planning process in the original DDQ \cite{peng2018integrating}. In DDQ, after each step of direct reinforcement learning, the agent improves its policy via $K$ steps of planning. A larger planning step means that more simulated experiences generated by $G$ are used for planning. Theoretically, larger amounts of \textit{high-quality} simulated experiences can boost the performance of the dialogue policy more quickly. However, the world model by no means perfectly reflects real human behavior, and the generated experiences, if of low quality, can have negative impact on dialogue policy learning. Prior work resorts to heuristics to mitigate the impact. For example, \citet{peng2018integrating} proposed to reduce planning steps at the late stage of policy learning, thus forcing all DDQ agents to converge to the same one trained with a small number of planning steps. 

\begin{figure}[t]
\centering
\includegraphics[width=1.0\linewidth]{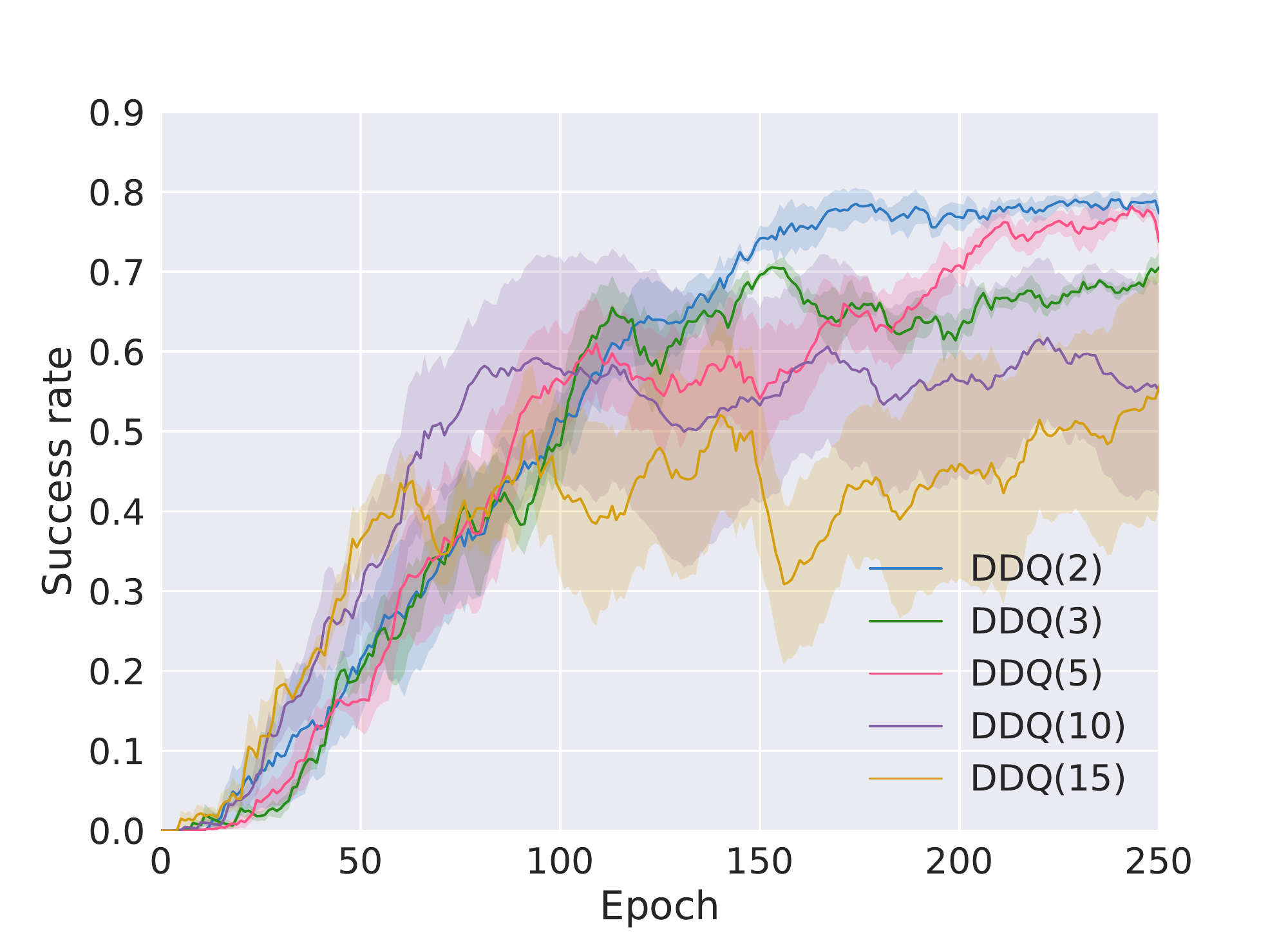}
\vspace{-5mm}
\caption{The learning curves of DDQ($K$) agents where $(K-1)$ is the number of planning steps.}
\label{fig:ddq_ps_2_3_5_10_15}
\end{figure}

Figure~\ref{fig:ddq_ps_2_3_5_10_15} shows the performance of DDQ agents with different planning steps without heuristics. It is observable that the performance is unstable, especially for larger planning steps, which indicates that the quality of simulated experience is becoming more pivotal as the number of planning steps increases.

D3Q resolves this issue by introducing a discriminator and allows only high-quality simulated experience, judged by the discriminator, to be used for planning. In the next section, we will show that D3Q does not suffer from the problem of DDQ and the D3Q training is quite stable even with large sizes of planning steps.

\section{Experiments}
We evaluate D3Q on the movie-ticket booking task with both simulated users and real users in two settings: full domain and domain extension.




\begin{table}[ht!]
\small
\begin{tabular}{|c|l|l|}
\hline
\multicolumn{2}{|c|}{Full Domain \& Domain Extension} \\
\hline
& \textsf{request}, \textsf{inform}, \textsf{deny}, \textsf{confirm\_question},\\ 
Intent & \textsf{confirm\_answer}, \textsf{greeting}, \textsf{closing}, \textsf{not\_sure},\\
 & \textsf{multiple\_choice}, \textsf{thanks}, \textsf{welcome} \\
\hline\hline
\multicolumn{2}{|c|}{Full Domain} \\
\hline
\multirow{4}{*}{Slot} & \textsf{city}, \textsf{closing}, \textsf{date}, \textsf{distanceconstraints}, \\
& \textsf{greeting}, \textsf{moviename}, \textsf{numberofpeople}, \\ 
& \textsf{price}, \textsf{starttime}, \textsf{state}, \textsf{taskcomplete}, \textsf{theater}, \\
& \textsf{theater\_chain}, \textsf{ticket}, \textsf{video\_format}, \textsf{zip} \\
\hline\hline
\multicolumn{2}{|c|}{Domain Extension} \\
\hline
\multirow{4}{*}{Slot} & \textsf{city}, \textsf{closing}, \textsf{date}, \textsf{distanceconstraints}, \\
& \textsf{greeting}, \textsf{moviename}, \textsf{numberofpeople}, \\ 
& \textsf{price}, \textsf{starttime}, \textsf{state}, \textsf{taskcomplete}, \textsf{theater}, \\
& \textsf{theater\_chain}, \textsf{ticket}, \textsf{video\_format}, \textsf{zip}, \\
& \textsf{genre}, \textsf{other}  \\
\hline
\end{tabular}
\centering
\caption{The data schema for full domain and domain extension settings.}
\label{tab:data_schema}
\end{table}


\subsection{Dataset}
\label{sec:dataset}
Raw conversational data in a movie-ticket booking scenario was collected via Amazon Mechanical Turk. The dataset has been manually labeled based on a schema defined by domain experts, as shown in Table~\ref{tab:data_schema}\nop{and~\ref{tab:data_schema_extension}}, consisting of 11 intents and 16 slots in the full domain setting, while there are 18 slots in the domain extension setting. Most of these slots can be both ``inform slots'' and ``request slots'', except for a few. For example, the slot \texttt{number\_of\_people} is categorized as an inform slot but not a request slot, because arguably the user always knows how many tickets she/he wants. In total, the dataset contains 280 annotated dialogues, the average length of which is approximately 11 turns.

\subsection{Baselines}
\label{sec:baselines}
To verify the effectiveness of D3Q, we developed different versions of task-completion dialogue agents as baselines to compare with.

\begin{itemize}
\item A \textbf{DQN} agent is implemented with only direct reinforcement learning in each episode.
\item The \textbf{DQN($K$)} has $K$ times more real experiences than the DQN agent.
The performance of DQN($K$) can be viewed as the upper bound of DDQ($K$) and D3Q($K$) with the same number of planning steps ($K-1$), as these models have the same training settings and the same amount of training samples during the entire learning process.
\item The \textbf{DDQ($K$)} agents are learned with an initial world model pre-trained on human conversational data, with $(K-1)$ as the number of planning steps. These agents store the simulated experience without being judged by the discriminator.
\end{itemize}

\paragraph{Proposed D3Q}
\begin{itemize}
\item The \textbf{D3Q($K$)} agents are learned through the process described in Section~\ref{subsec:controlled_planning}.
\item The \textbf{D3Q($K$, fixed $\theta_D$)} agents are learned as described in Section~\ref{subsec:controlled_planning} without training discriminator. The D3Q($K$, fixed $\theta_D$) agents are only evaluated in the simulation setting.
\end{itemize}

\subsection{Implementation}
\label{sec: implementation}
\paragraph{Settings and Hyper-parameters}
$\epsilon$-greedy is always applied for exploration. We set the discount factor $\gamma$ = 0.9. The buffer size of $B^u$ and $B^h$ is set to 2000 and 2000 $\times K$ \emph{planning steps}, respectively. The batch size is 16, and the learning rate is 0.001. 
To prevent gradient explosion, we applied gradient clipping on all the model parameters to maximum norm = 1. All the NN models are randomly initialized. The high-quality simulated experience buffer $B^h$ and the simulated experience buffer $B^s$ are initialized as empty. The target network is updated at the beginning of each training episode. The optimizer for all the neural networks is RMSProp~\cite{hinton2012neural}. The maximum length of a simulated dialogue is 40. If exceeding the maximum length, the dialogue fails. To make dialogue training efficient, we also applied a variant of imitation learning, called Reply Buffer Spiking (RBS)~\cite{lipton2016efficient}, by building a simple and straightforward rule-based agent based on human conversational dataset. We then pre-filled the real experience replay buffer $B^u$ with experiences of 50 dialogues, before training for all the variants of models. The batch size for collecting experiences is 10, which means if the running agent is DDQ/D3Q($K$), 10 real experience tuples and 10 $\times$ ($K-1$) simulated experience tuples are stored into the buffers at every episode.

\paragraph{Agents}
For all the models (DQN, DDQ, and D3Q) and their variants, the value networks $Q(.)$ are MLPs with one hidden layer of size 80 and ReLU activation.

\paragraph{World Model}
For all the models (DDQ and D3Q) and their variants, the world models $M(.)$ are MLPs with one shared hidden layer of size 160, hyperbolic-tangent activation, and one encoding layer of hidden size 80 for each state and action input.

\paragraph{Discriminator} 
In the proposed D3Q framework, the LSTM cell is utilized, the hidden size is 128. The encoding layer for the current state and output layer are MLPs with single hidden layer of size 80. The threshold interval is set to range between 0.45 and 0.55, i.e., only when $0.45 \leq D(x) \leq 0.55$ that $x$ would be stored into the buffer $B^h$.

\begin{table*}[t]
\small
\begin{center}
\begin{tabular}{lrrrrrrrrr}
\\ \hline
\multirow{2}{*}{\bf Agent}& \multicolumn{3}{c}{\bf Epoch = 100} & \multicolumn{3}{c}{\bf Epoch = 200} & \multicolumn{3}{c}{\bf Epoch = 300} \\ 
\cline{2-10}
 & \bf Success & \bf Reward & \bf Turns & \bf Success & \bf Reward & \bf Turns & \bf Success & \bf Reward & \bf Turns \\ \hline
DQN & .4467 & 2.993 & 23.21  & .7000 & 36.08 & 17.84 & .7867 & 48.45 & 13.91 \\ 
\hline
DDQ(5) & .5467 & 16.57 & 20.07 & \bf .7133 & \bf 39.23 & \bf 14.73 & \bf .8067 & \bf 50.73 & 14.13 \\
DDQ(5, rand-init $\theta_G$) & .6067 & 23.55 & 20.49 & .6267 & 26.30 & 19.80 & .6667 & 32.92 & 16.16\\ 
DDQ(5, fixed $\theta_G$) & .5867 & 20.62 & 21.56 & .1667 & -33.71 & 29.41 & .2267 & -22.68 & 21.76\\ 
D3Q(5) & \bf .7467 & \bf 43.59 & \bf 14.03 & .6800 & 34.64 & 15.92 & .7200 & 40.85 & \bf 13.11\\
D3Q(5, fixed  $\theta_D$) & .6800 & 33.86 & 17.48 &  .7000 & 36.57 & 16.85 & .6933 & 35.67 & 17.06 \\ 
DQN(5) & \textit{.7400} & \textit{42.19} & \textit{15.23} & \textit{.8533} & \textit{57.76} & \textit{11.28} & \textit{.7667} & \textit{46.56} & \textit{12.88} \\ 
\hline
DDQ(10) & .5733 & 24.00 & \bf 11.60 & .5533 & 19.89 & 15.01 & .4800 & 10.04 & 17.12\\
DDQ(10, rand-init $\theta_G$) & .5000 & 12.79 & 16.41 & .5333 & 17.71 & \bf 14.57 & .6000 & 24.98 & 16.04\\ 
DDQ(10, fixed $\theta_G$) & .3467 & -10.25 & 25.69 & .2400 & -23.38 & 26.36 & .0000 & -55.53 & 33.07\\
D3Q(10) & .6333 & 28.99 & 16.01 & .7000 & 37.24 & 15.52 & .6667 & 33.09 & 15.83\\ 
D3Q(10, fixed  $\theta_D$) & \bf .7133 & \bf 36.36 & 20.48 & \bf .8400 & \bf 54.87 & 20.48 & \bf .7400 & \bf 42.89 & \bf 13.81\\
DQN(10) & \textit{.8333} & \textit{55.5} & \textit{11.00} & \textit{.7733} & \textit{47.99} & \textit{11.61} & \textit{.7733} & \textit{47.68} & \textit{12.24} \\
\hline
\end{tabular}
\end{center}
\vspace{-2mm}
\caption{Results of different agents at training epoch = \{100, 200, 300\}. Each number is averaged over 3 runs, each run tested on 50 dialogues. (Success: success rate, Reward: Average Reward, Turns: Average Turns)}
\label{tab:test_results}
\vspace{-2mm}
\end{table*}

\subsection{Simulation Evaluation}
\label{subsec:simulation_evaluation}
In this setting, the dialogue agents are optimized by interacting with the user simulators instead of with real users. In another word, the world model is trained to mimic user simulators. In spite of the discrepancy between simulators and real users, this setting endows us with the flexibility to perform a detailed analysis of models without much cost, and to reproduce experimental results easily.

\paragraph{User Simulator} 
We used an open-sourced task-oriented user simulator~\cite{li2016user} in our simulated evaluation experiments (Appendix~\ref{app:user_sim} for more details). The simulator provides the agent with a simulated user response in each dialogue turn along with a reward signal at the end of the dialogue. A dialogue is considered successful if and only if a movie ticket is booked successfully, and the information provided by the agent satisfies all the constraints of the sampled user goal. At the end of each dialogue, the agent receives a positive reward $2*L$ for success, or a negative reward $-L$ for failure, where $L$ is the maximum number of turns in each dialogue, and is set to $40$ in our experiments. Furthermore, in each turn, a reward $-1$ is provided to encourage shorter dialogues.

\begin{figure}[t]
\centering
\includegraphics[width=1.0\linewidth]{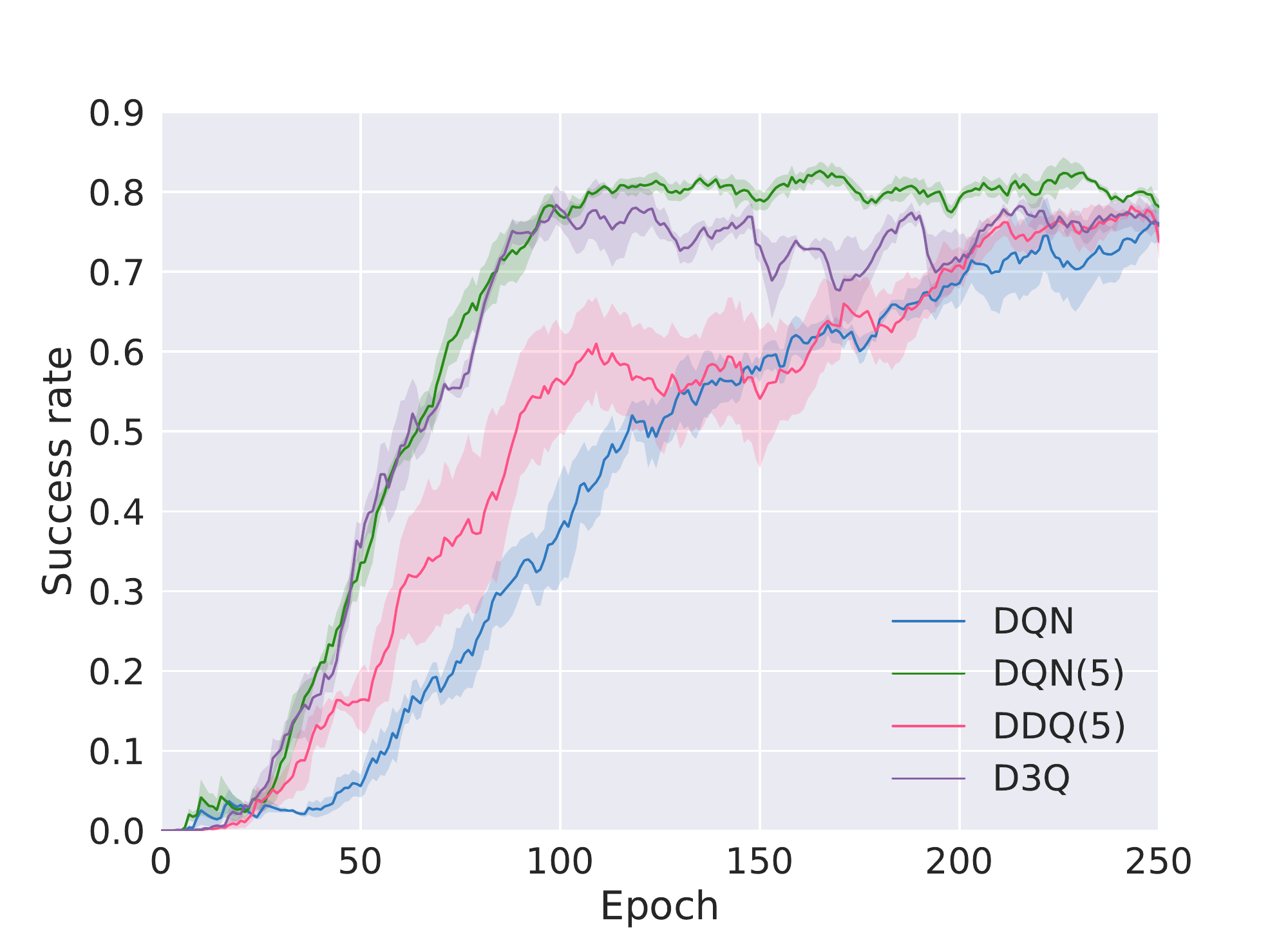}
\vspace{-5mm}
\caption{The learning curves of agents (DQN, DDQ, and D3Q) under the full domain setting.}
\label{fig:dqn_ddq_d3q}
\end{figure}

\begin{figure}[t]
\centering
\includegraphics[width=1.0\linewidth]{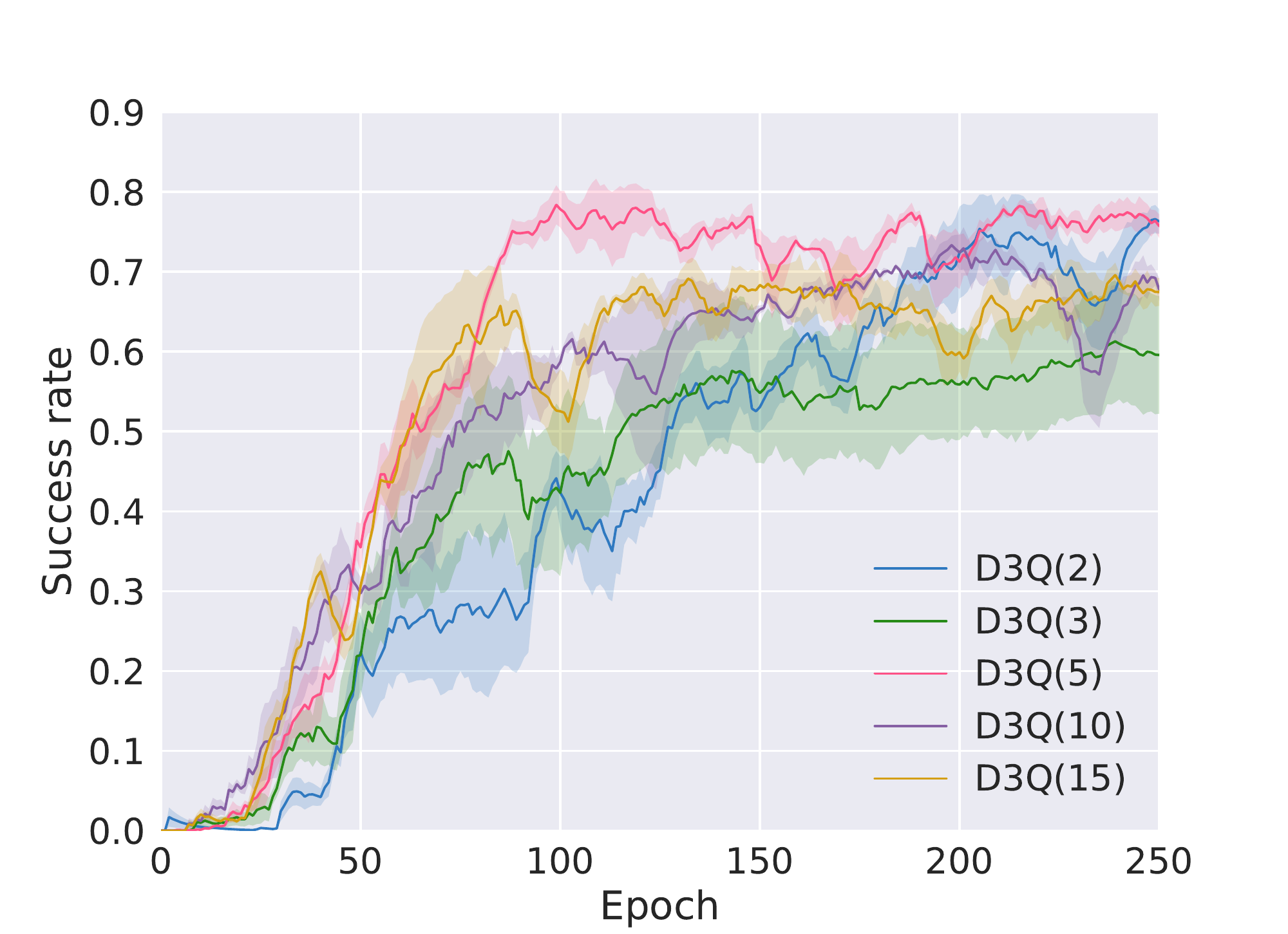}
\vspace{-5mm}
\caption{The learning curves of D3Q(K) agents which (K-1) is the number of planning steps (K = 2, 3, 5, 10, 15).}
\label{fig:d3q_ps_2_3_5_10_15}
\end{figure}

\begin{figure}[t]
\centering
\includegraphics[width=1.0\linewidth]{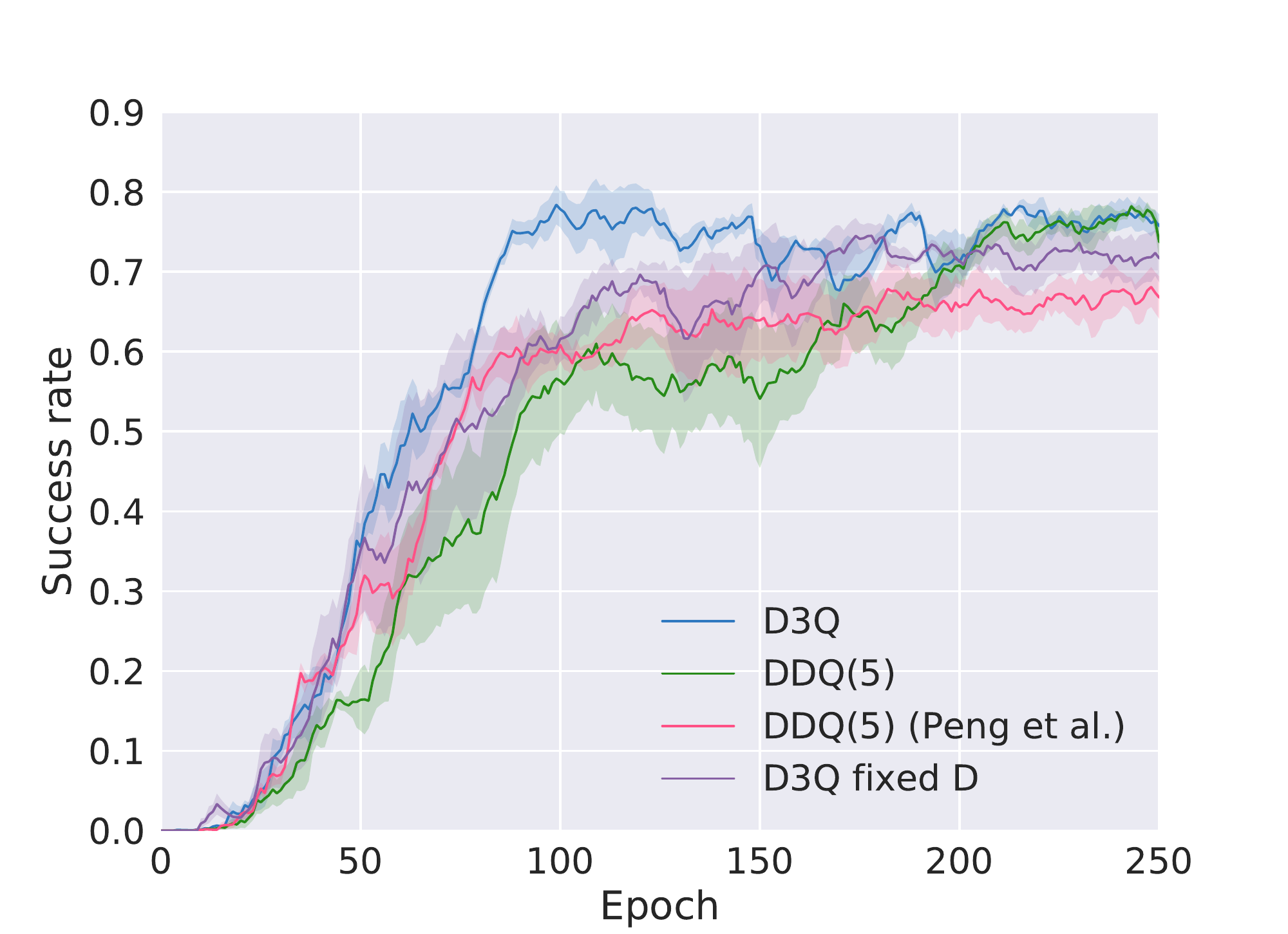}
\vspace{-5mm}
\caption{The learning curves of D3Q, DDQ(5), DDQ(5)~\cite{peng2018integrating}, and D3Q fixed $\theta_D$ agents.}
\label{fig:d3q_ddq_ddq_old_d3q_fixedD}
\end{figure}

\paragraph{Full Domain}
The learning curves of the models in the full domain setting are depicted in the figure~\ref{fig:dqn_ddq_d3q}. The results show that the proposed D3Q agent (the pink curve) significantly outperforms the baselines DQN and DDQ(5), and has similar training efficiency to DQN(5). Note that here the planning steps of D3Q is 4, which means D3Q (pink) and DDQ(5) (purple) use the same amount of training samples (both real and simulated experiences) to update the agent throughout the whole training process. The difference between these two agents is that D3Q employs a discriminator as a quality judge. The experimental result shows that our proposed framework could boost the learning efficiency even without any pre-training on the discriminator. Furthermore, D3Q (pink) uses the same amount of training samples as DQN(5) (green), while the proposed model uses only 20\% of real experience from human. The efficacy and feasibility of D3Q is hereby justly verified.

As mentioned in the previous section, a large number of planning steps means leveraging a large amount of simulated experience to train the agents. The experimental result (Figure~\ref{fig:ddq_ps_2_3_5_10_15}) shows that the DDQ agents are highly sensitive to the quality of simulated experience.
In contrast, the proposed D3Q framework demonstrates robustness to the number of planning steps (Figure~\ref{fig:d3q_ps_2_3_5_10_15}). Figure~\ref{fig:d3q_ddq_ddq_old_d3q_fixedD} shows that D3Q also outperforms DDQ original setting ~\cite{peng2018integrating} and D3Q without training discriminator.
The performance detail including success rate, reward, an number of turns is shown in Table~\ref{tab:test_results}. 
From the table, with fewer simulated experiences, the difference between DDQ and D3Q may not be significant, where DDQ agents achieve about 50\%-60\% success rate and D3Q agents achieve higher than 68\% success rate after 100 epochs.
However, when the number of planning steps increases, more fake experiences significantly degrade the performance for DDQ agents, where DDQ(10, fixed $\theta_G$) suffers from bad simulated experiences after 300 epochs and achieves $0\%$ success rate.

\begin{figure}[t]
\centering
\includegraphics[width=\linewidth]{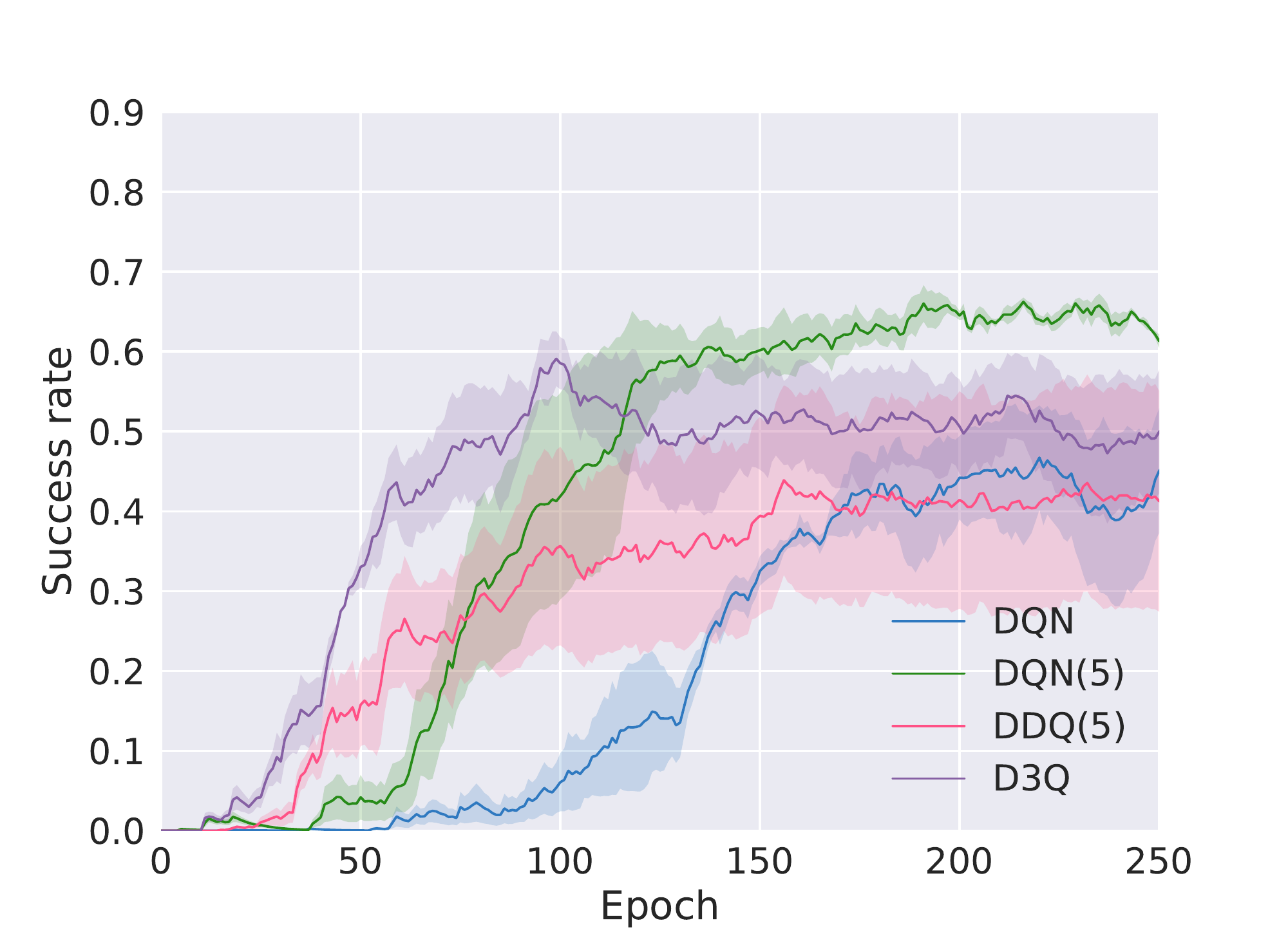}
\vspace{-5mm}
\caption{The learning curves of agents (DQN, DDQ, and D3Q) under the domain extension setting.}
\label{fig:de}
\end{figure}

\paragraph{Domain Extension}
In the domain extension experiments, more complicated user goals are adopted. Moreover, we narrow down the action space into a small subspace instead of that used in full-domain setting, and gradually introduce more complex user goals and expand the action space as the training proceeds. Specifically, we start from a set of necessary slots and actions to accomplish most of the user goals, and then extend the action space and complexity of user goals once every 20 epoch (after epoch 50). Note that the domain will keep extending and become full-domain after epoch 130. Such experimental setting makes the training environment more complicated and unstable than the previous full-domain one.

The results summarized in Figure~\ref{fig:de} show that D3Q significantly outperforms the baseline methods, demonstrating its robustness. Furthermore, D3Q shows remarkable learning efficiency while extending the domain, which even outperforms DQN(5). A potential reason might be that the world model could improve exploration in such unstable and noisy environment.

\subsection{Human Evaluation}
In the human evaluation experiments, real users interact with different models without knowing which agent is behind the system. At the beginning of each dialogue session, one of the agents was randomly picked to converse with the user. The user was instructed to converse with the agent to complete a task given a user goal sampled from the corpus. The user can abandon the task and terminate the dialogue at any time, if she or he believes that the dialogue was unlikely to succeed, or simply because the dialogue drags on for too many turns. In such cases, the dialogue session is considered as failure.

\begin{figure}[t]
\centering
\includegraphics[width=1.0\linewidth]{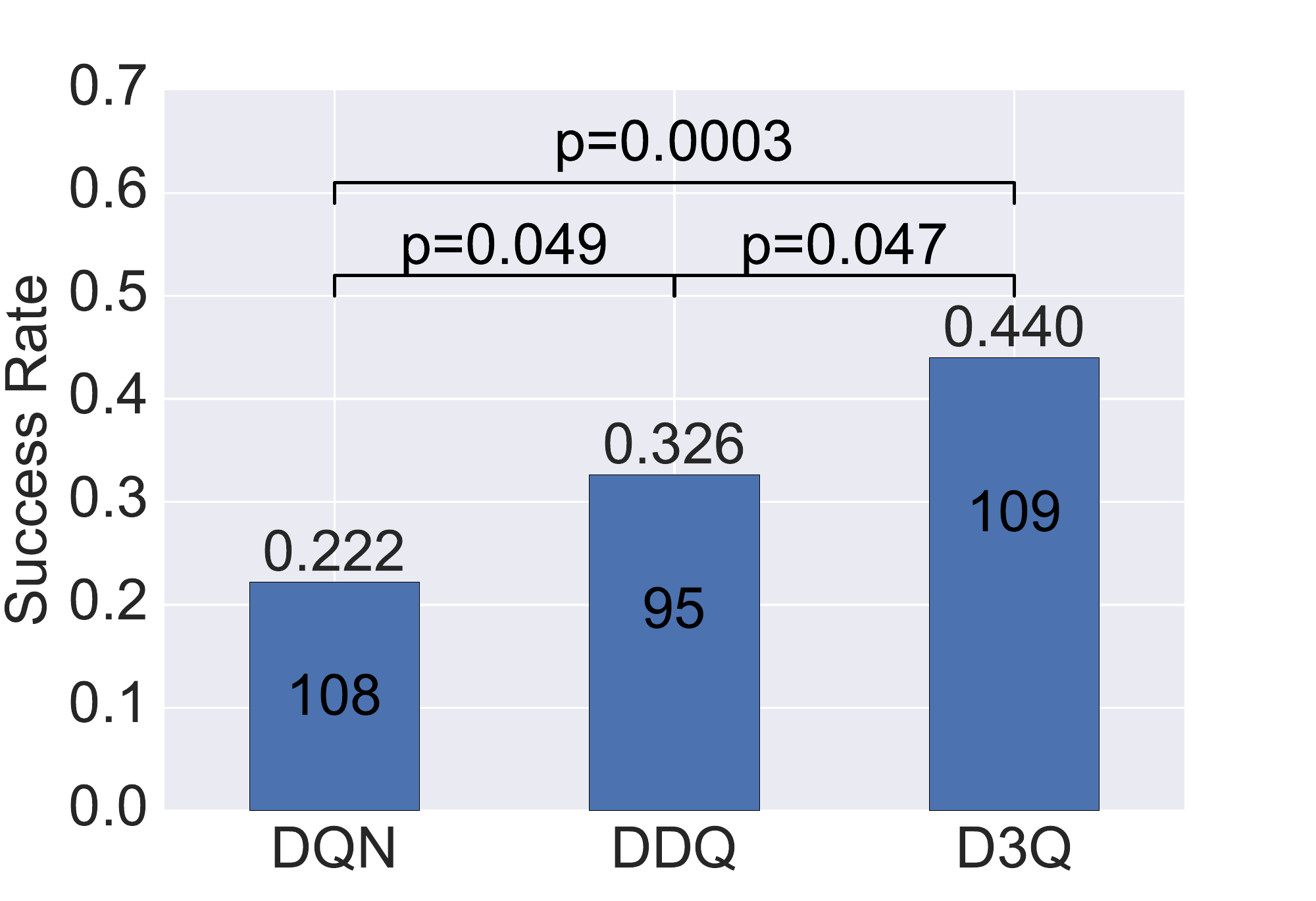}
\vspace{-3mm}
\caption{The human evaluation results of D3Q, DDQ(5), and D3Q in the full domain setting, the number of test dialogues indicated on
each bar, and the p-values from a two-sided permutation test (difference in mean is significant with $p<0.05$).}
\label{fig:he_full}
\end{figure}

\begin{figure}[t!]
\centering
\includegraphics[width=1.0\linewidth]{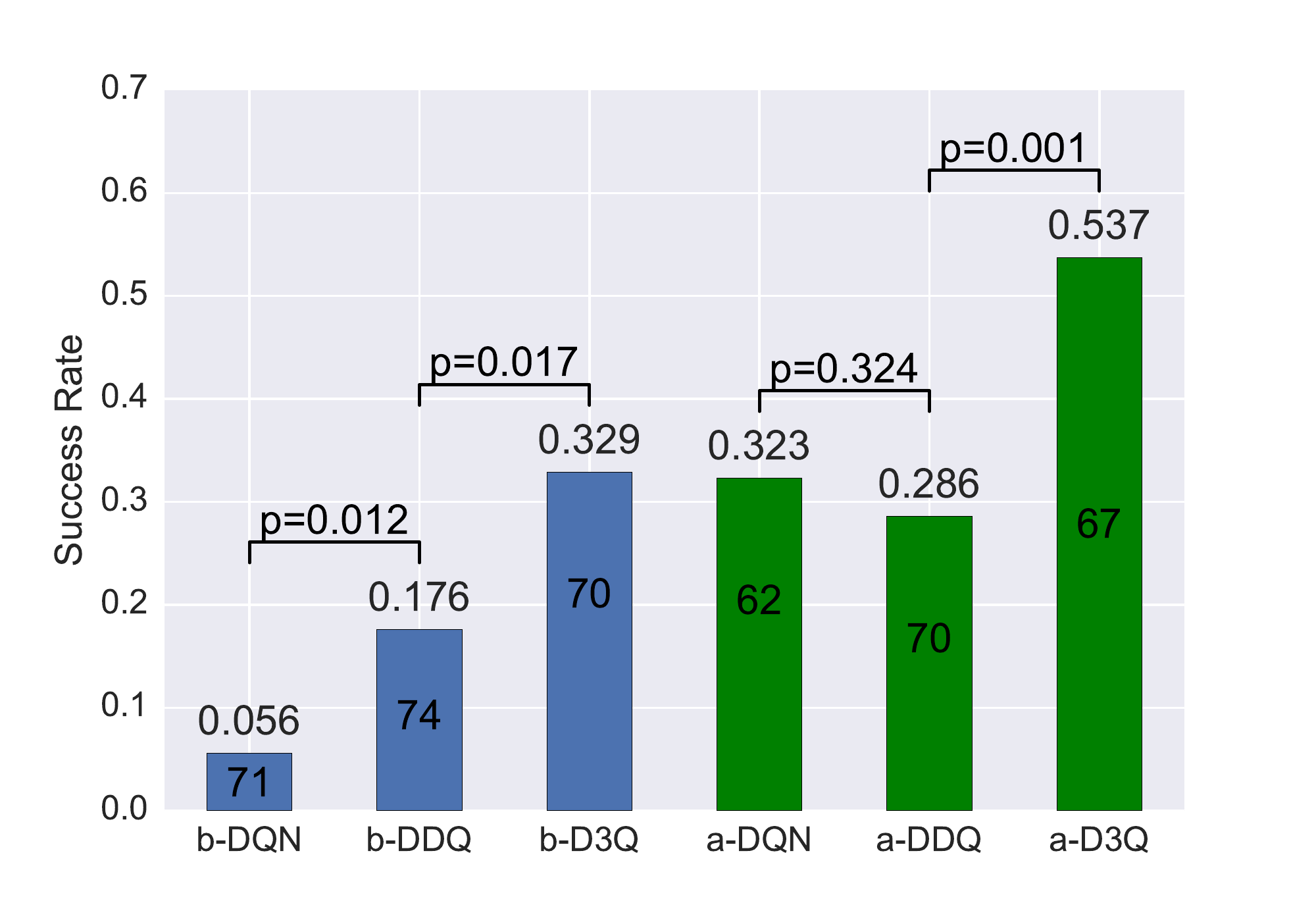}
\vspace{-3mm}
\caption{The human evaluation results of DQN, DDQ(5), and D3Q in the domain extension setting, the number of test dialogues indicated on each bar. The prefix 'b-' implies that the trained models are picked before the environment extends to full domain, while the prefix 'a-' indicates that the trained models are picked after the environment becomes full domain (difference in mean is significant with $p<0.05$).}
\label{fig:he_de}
\end{figure}

\paragraph{Full Domain}
Three agents (DQN, DDQ(5), and D3Q) trained in the full domain setting (Figure~\ref{fig:dqn_ddq_d3q}) at epoch 100 are selected for testing. As illustrated in Figure~\ref{fig:he_full}, the results of human evaluation are consistent with those in the simulation evaluation (Section~\ref{subsec:simulation_evaluation}), and the proposed D3Q significantly outperforms other agents.

\paragraph{Domain Extension}
To test the adaptation capability of the agents to the complicated, dynamically changing environment, we selected three trained agents (DQN, DDQ(5), and D3Q) at epoch 100 before the environment extends to full domain, and another three agents trained at epoch 200 after the environment becomes full domain. Figure~\ref{fig:he_de} shows that the results are consistent with those in the simulation evaluation (Figure~\ref{fig:de}), and the proposed D3Q significantly outperforms other agents in both stages.

\section{Conclusions}
This paper proposes a new framework, Discriminative Deep Dyna-Q (D3Q), for task-completion dialogue policy learning. With a discriminator as judge, the proposed approach is capable of controlling the quality of simulated experience generated in the planning phase, which enables efficient and robust dialogue policy learning. Furthermore, D3Q can be viewed as a generic model-based RL approach easily-extensible to other RL problems. 

We validate the D3Q-trained dialogue agent on a movie-ticket-booking task in the simulation, human evaluation, and domain-extension settings. Our results show that the D3Q agent significantly outperforms the agents trained using other state-of-the-art methods including DQN and DDQ.

\section*{Acknowledgments}
We thank the anonymous reviewers for their insightful feedback on the work. We would like to acknowledge the volunteers from Microsoft for participating the human evaluation experiments. 
Shang-Yu Su and Yun-Nung Chen are supported by the Ministry of Science and Technology of Taiwan and MediaTek Inc.

\bibliography{emnlp2018}
\bibliographystyle{acl_natbib_nourl}

\appendix

\section{User Simulator}
\label{app:user_sim}
In the task-completion dialogue setting, the entire conversation is around a user goal implicitly, but the agent knows nothing about the user goal explicitly and its objective is to help the user to accomplish this goal. Generally, the definition of user goal contains two parts: 

\begin{compactitem}
\item \emph{inform\_slots} contain a number of slot-value pairs which serve as constraints from the user.
\item \emph{request\_slots} contain a set of slots that user has no information about the values, but wants to get the values from the agent during the conversation. \textsf{ticket} is a default slot which always appears in the \emph{request\_slots} part of user goal.
\end{compactitem}

To make the user goal more realistic, we add some constraints in the user goal: slots are split into two groups. Some of slots must appear in the user goal, we called these elements as \emph{Required slots}. In the movie-booking scenario, it includes \textsf{moviename, theater, starttime, date, numberofpeople}; the rest slots are \emph{Optional slots}, for example, \textsf{theater\_chain, video\_format} etc.

We generated the user goals from the labeled dataset using two mechanisms. One mechanism is to extract all the slots (known and unknown) from the first user turns (excluding the greeting user turn) in the data, since usually the first turn contains some or all the required information from user. The other mechanism is to extract all the slots (known and unknown) that first appear in all the user turns, and then aggregate them into one user goal. We dump these user goals into a file as the user-goal database. Every time when running a dialogue, we randomly sample one user goal from this user goal database.

\end{document}